\documentclass[letterpaper]{article}
\usepackage[preprint]{aaai2027}
\usepackage[hyphens]{url}
\usepackage{graphicx}
\usepackage{natbib}
\usepackage{caption}
\usepackage{amsmath}
\usepackage{amssymb}
\usepackage{booktabs}
\usepackage{multirow}
\usepackage{cleveref}

\title{RISE: Roadside Infrastructure Sequence Understanding across 3D Tracking and Structured Vision-Language Reasoning}

\author{
Yanbo Jiang\textsuperscript{\rm 1},
Haotian Zheng\textsuperscript{\rm 1},
Jiahao Wang\textsuperscript{\rm 1},
Hanxiao Ren\textsuperscript{\rm 1},
Yitao Xu\textsuperscript{\rm 1},
Yining Xing\textsuperscript{\rm 1},
Zehong Ke\textsuperscript{\rm 1},
Hao Cheng\textsuperscript{\rm 1},
Yiqian Tu\textsuperscript{\rm 1},
Jinhao Li\textsuperscript{\rm 1},
Zhiyuan Xuan\textsuperscript{\rm 2},
Fang Zhang\textsuperscript{\rm 1}\protect\footnotemark[1],
Jianqiang Wang\textsuperscript{\rm 1}\protect\thanks{Corresponding authors: \{zhang\_fang,wjqlws\}@tsinghua.edu.cn}
}
\affiliations{
\textsuperscript{\rm 1}School of Vehicle and Mobility, Tsinghua University
\quad
\textsuperscript{\rm 2}TsingCloud
}

\begin{document}
\maketitle

\begin{abstract}
We present RISE (\textbf{R}oadside \textbf{I}nfrastructure \textbf{S}equence Understanding and \textbf{E}valuation), a framework spanning metric 3D tracking and structured vision-language reasoning in roadside sequences.
For metric tracking, our image-only method combines SAM3 video identities with calibration-guided mask agreement for multi-view identity association, recovering persistent 3D tracks without LiDAR or task-specific 3D training.
Its calibration-conditioned geometry allows the procedure to be instantiated at different calibrated multi-camera intersections without layout-specific retraining.
On 20 human-reviewed clips from six intersections, the generated tracks achieve 66.9 MOTA within the defined multi-view evaluation scope.
For structured vision-language reasoning, a human-reviewed MLLM pipeline mines high-value clips and uses a constrained full-context Oracle to construct bbox-grounded predictive QA without exposing future evidence to evaluated models.
The resulting RISE-VQA dataset contains 33,910 QA pairs from 557 clips across 16 intersections and 61 roadside views.
Its intersection-held-out RISE-Bench evaluates semantic choices, coordinates, future boxes, and interaction sets with deterministic task-specific metrics.
Experiments show consistent benefits from domain adaptation and generally from temporal context, while revealing persistent challenges in spatial grounding, future localization, and interaction reasoning.
\end{abstract}

\section{Introduction}

Fixed roadside cameras are a key sensing layer for intelligent transportation systems, supporting intersection monitoring, traffic-safety analysis, and infrastructure-assisted perception through persistent, scene-centric observations~\cite{yu2023v2x}.
Turning these sequences into actionable traffic understanding requires two complementary capabilities: metric 3D tracking establishes persistent identities, 3D states, and trajectories, while structured vision-language reasoning provides queryable interpretations of agent states, maneuvers, interactions, and future outcomes.
Prior work has largely studied these capabilities separately, emphasizing metric tracking and trajectory forecasting in vehicle--infrastructure sequences~\cite{yu2023v2x} or language-based traffic-scene interpretation in driving VQA~\cite{xu2021sutd,qian2024nuscenes,sima2024drivelm}.

These capabilities draw on complementary spatial and temporal evidence in fixed roadside sequences.
Spatially, synchronized cameras with overlapping fields of view provide multi-view geometric constraints for metric 3D estimation; temporally, complete sequences capture event evolution and outcomes needed for predictive reasoning.
Yet converting this information into reusable supervision remains challenging.
Camera-only roadside 3D detection often requires task-specific 3D labels or adaptation to new camera layouts~\cite{yang2023bevheight,yang2024monogae,zhang2025mic}.
For predictive VQA, high-value events are sparse, and realized outcomes must provide supervision without exposing future evidence to evaluated models.
Scaling roadside collection across intersections further requires access, calibration, and synchronization at each site.

To address these challenges, we introduce RISE (\textbf{R}oadside \textbf{I}nfrastructure \textbf{S}equence Understanding and \textbf{E}valuation), which exploits complementary spatial and temporal evidence in fixed-camera roadside sequences.
Its tracking branch combines calibrated multi-view agreement with video identities to construct persistent metric 3D tracks, while its vision-language branch converts realized event outcomes into bbox-grounded predictive supervision through a constrained Oracle.
Together, the two branches organize roadside sequence understanding around persistent traffic agents, covering metric state estimation and semantic event interpretation with task-specific inputs.

Our contributions are:
\begin{enumerate}
    \item \textbf{A roadside sequence task framework.}
    We formulate metric 3D tracking and structured vision-language reasoning as complementary tasks centered on persistent traffic agents while retaining task-specific inputs.

    \item \textbf{Training-free multi-view 3D tracking.}
    We combine SAM3 view-local identities with calibration-guided mask agreement, MWIS-based association, and track-level refinement to recover persistent metric 3D tracks without LiDAR or task-specific 3D training.
    The calibration-conditioned procedure can be instantiated at different calibrated multi-camera intersections, and its outputs initialize subsequent human refinement and quality assessment.

    \item \textbf{RISE-VQA with Oracle-grounded predictive supervision.}
    We construct 33,910 bbox-grounded QA pairs from 557 clips across 16 intersections and 61 roadside views.
    A human-reviewed MLLM pipeline uses a constrained full-context Oracle to generate predictive targets without exposing future evidence to evaluated models.

    \item \textbf{Intersection-held-out structured evaluation.}
    We introduce RISE-Bench to evaluate semantic choices, coordinates, future boxes, and interaction sets on unseen intersections using deterministic task-specific metrics.
    Baselines quantify domain-adaptation and temporal-context effects while exposing limitations in grounding, prediction, and interaction reasoning.
\end{enumerate}

\section{Related Work}

\subsection{Roadside 3D Perception and Tracking}

Roadside perception datasets such as DAIR-V2X~\cite{yu2022dair}, V2X-Seq~\cite{yu2023v2x}, RCooper~\cite{hao2024rcooper}, and Rope3D~\cite{ye2022rope3d} provide calibrated infrastructure sensors and 3D annotations for vehicle-infrastructure perception.
V2X-Seq further augments sequential frames with persistent identities, trajectories, vector maps, and traffic-light signals, and defines VIC3D tracking together with online and offline forecasting~\cite{yu2023v2x}.
I-24 3D provides continuous vehicle trajectories from overlapping calibrated highway cameras~\cite{gloudemans2023i24}.
RISE shares this sequence-centric perspective but studies a different setting: fixed multi-camera, image-only 3D tracking together with structured vision-language reasoning over roadside events.

Camera-only roadside 3D methods exploit camera calibration, ground geometry, or BEV fusion~\cite{yang2023bevheight,yang2024monogae,zhang2025mic}, but generally learn 3D perception from task-specific supervision.
In contrast, our method uses video-consistent segmentation identities as temporal evidence and calibration-guided mask agreement as cross-view geometric evidence.
Its projection geometry is instantiated directly from each deployment's calibration, allowing the same procedure to operate at a new calibrated multi-camera intersection with sufficient overlap, without training a layout-specific 3D model.

\subsection{Structured Vision-Language Reasoning for Roadside Scenes}

\begin{table*}[!t]
\centering
\small
\caption{Comparison with traffic-scene vision-language datasets. ``Inherited'' denotes 3D labels from a source dataset; ``Img. tracker'' denotes the image-derived tracking component proposed in RISE.}
\label{tab:dataset_comparison}
\setlength{\tabcolsep}{1.6pt}
\begin{tabular}{@{}lcccccccc@{}}
\toprule
\textbf{Dataset} & \textbf{Venue} & \textbf{View} & \textbf{Unit} & \textbf{Source} & \textbf{QA Gen.} & \textbf{\#QA} & \textbf{Scoring} & \textbf{3D Comp.} \\
\midrule
NuScenes-QA~\cite{qian2024nuscenes} & AAAI'24 & ego & Multi-view & nuScenes & Template & 460k & Accuracy & Inherited \\
DriveLM~\cite{sima2024drivelm} & ECCV'24 & ego & Frame & nuScenes & Temp.+Human & 443k & SPICE/GPT & Inherited \\
LingoQA~\cite{marcu2024lingoqa} & ECCV'24 & ego & Video & Self & Human+LLM & 420k & Lingo-Judge & -- \\
SUTD-TrafficQA~\cite{xu2021sutd} & CVPR'21 & mixed & Video & Mixed & Human & 62.5k & Acc. & -- \\
\midrule
TUMTraf VideoQA~\cite{zhou2025tumtraf} & ICML'25 & roadside & Video & Self & LLM+Ver. & 85.0k & Acc. & -- \\
RoadSceneVQA~\cite{guan2026roadscenevqa} & AAAI'26 & roadside & Image & Rope3D & LLM+Ver. & 34.7k & Text/GPT & Inherited \\
V2X-QA~\cite{you2026v2x} & arXiv'26 & V2X & View pair & V2X-Seq & LLM+Ver. & 33.2k & Acc. & Inherited \\
LTD~\cite{huang2026towards} & arXiv'26 & roadside & Multi-img & Self & LLM+Ver. & 11.6k & GPT/Acc. & -- \\
\midrule
\textbf{RISE (Ours)} & -- & \textbf{roadside} & \textbf{Clip} & \textbf{Self} & \textbf{LLM+Ver.} & \textbf{33.9k} & \textbf{Task-specific} & \textbf{Img. tracker} \\
\bottomrule
\end{tabular}
\end{table*}

Driving-oriented VQA datasets include SUTD-TrafficQA~\cite{xu2021sutd}, NuScenes-QA~\cite{qian2024nuscenes}, DriveLM~\cite{sima2024drivelm}, and Talk2BEV~\cite{choudhary2024talk2bev}.
NuScenes-QA and DriveLM construct QA from vehicle-centric autonomous-driving data, while SUTD-TrafficQA targets video reasoning over mixed traffic events.
Recent work extends this scope to infrastructure and cooperative settings: RoadSceneVQA~\cite{guan2026roadscenevqa} studies image-level roadside reasoning, TUMTraf VideoQA~\cite{zhou2025tumtraf} targets spatiotemporal roadside video understanding, V2X-QA~\cite{you2026v2x} evaluates cooperative VQA on V2X-Seq, and LTD/UniVLT~\cite{huang2026towards} considers heterogeneous multi-camera reasoning tasks.

As summarized in \Cref{tab:dataset_comparison}, RISE complements existing benchmarks in temporal input, target grounding, output structure, and evaluation protocol.
It uses newly collected clips from fixed roadside cameras to capture short-term traffic evolution while keeping the visual context compact.
Each queried road user is referenced by its observed bounding box rather than a descriptive phrase, requiring the model to visually ground and associate the agent across frames.
The answers include semantic choices, spatial coordinates, future 2D bounding boxes, and set-valued interactions.
RISE-Bench scores these heterogeneous outputs with deterministic task-specific metrics and holds out complete intersections to evaluate generalization to unseen intersection layouts.

\section{Method}

\subsection{Sequence Task Formulation}

Let $I_t^{(v)}$ denote frame $t$ from fixed camera $v$.
Given a completed roadside recording, the 3D tracking branch uses four synchronized views to recover
\begin{equation}
\mathcal{T}=\{\tau_i\},\qquad
\tau_i=\{(\mathbf b^{3D}_{i,t},\mathrm{id}_i)\}_{t},
\end{equation}
where $\mathbf b^{3D}_{i,t}$ contains the metric center, dimensions, and heading of agent $i$, and $\mathrm{id}_i$ identifies the agent across the sequence.

The VQA branch instead evaluates reasoning under partial observation.
Each view is processed independently to avoid the visual-token and cross-view association burden of multi-view temporal input.
For each 20-frame clip, the evaluated VLM observes only $I_0^{(v)}$--$I_4^{(v)}$, whereas the annotation Oracle may inspect the complete sequence $I_0^{(v)}$--$I_{19}^{(v)}$ solely to determine supervision targets.
Each QA instance grounds a queried agent with an observed 2D box and requires a semantic choice, spatial coordinates, future 2D boxes, or an interaction set.
The two tasks use roadside sequences differently: 3D tracking exploits multi-view temporal identity evidence over a completed recording, whereas VQA evaluates an observed prefix using supervision derived from the complete event.
Together, they form RISE's sequence-understanding framework.

\subsection{Identity-Backbone Multi-View 3D Tracking}

SAM3 provides temporally consistent mask IDs within each camera, but these IDs are not shared across views.
RISE converts them into calibration-conditioned cross-view signatures and uses MWIS to select non-conflicting identity backbones, establishing global object identities before metric cuboid fitting.
Unlike frame-level 3D detect-then-associate pipelines, this identity-first design turns SAM3's temporal continuity into multi-view 3D tracks and retains association evidence when some views are occluded.

\begin{figure}[!t]
    \centering
    \includegraphics[width=\columnwidth]{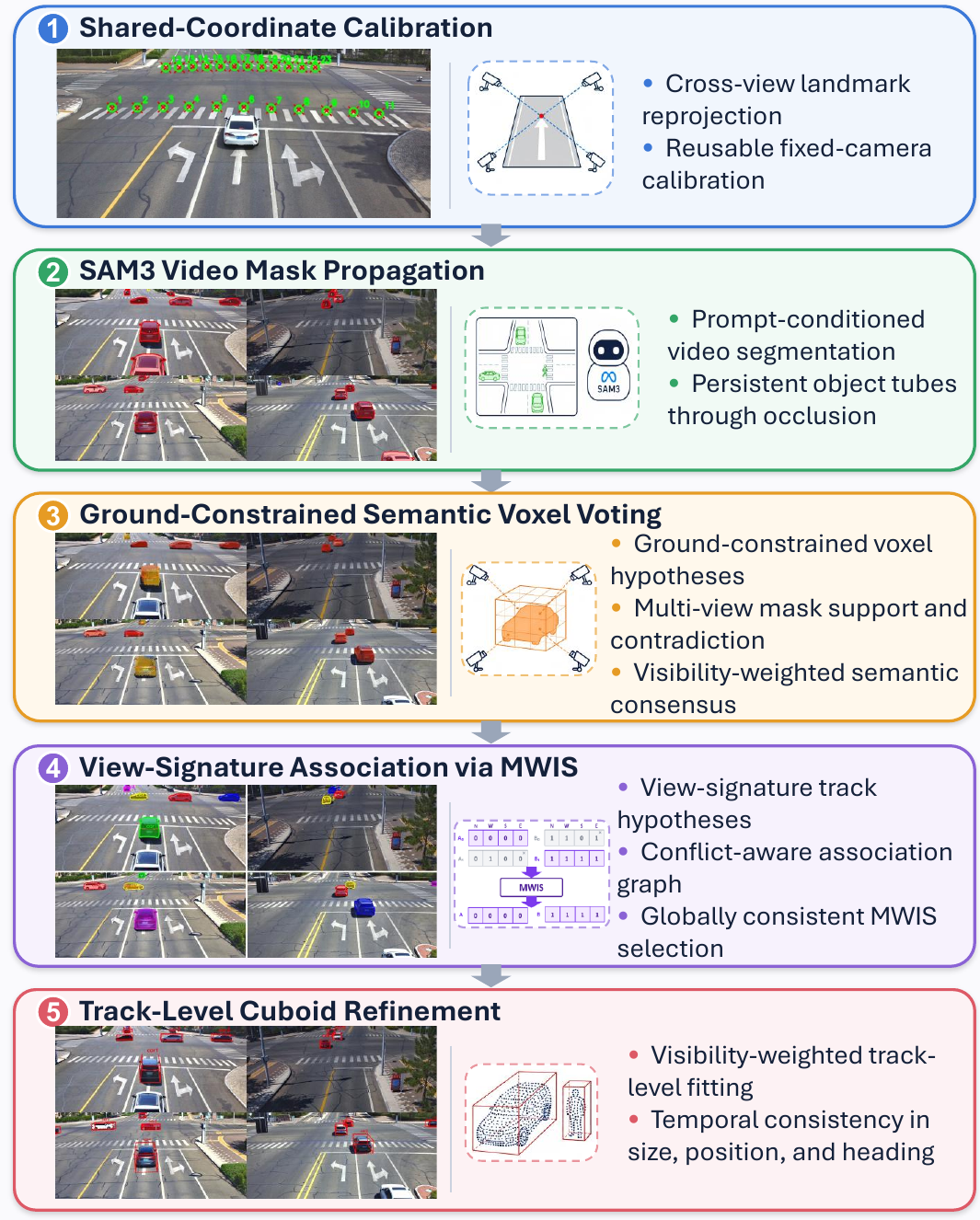}
    \caption{Identity-first multi-view 3D tracking in RISE. View-local SAM3 video IDs are grouped into global identity backbones through calibrated voxel signatures and MWIS, followed by metric cuboid fitting and track-level refinement.}
    \label{fig:tracker}
\end{figure}

\subsubsection{Video IDs and Calibrated Semantic Voxels}

Given four synchronized and calibrated views, SAM3~\cite{carion2025sam} produces category-specific instance masks with temporally consistent IDs within each view.
We additionally mark static occluders to distinguish occlusion from missing segmentation.
We project a cached 3D voxel grid into each image and record the mask ID covering each projected voxel.
For voxel $n$ at time $t$, the cross-view tuple
\begin{equation}
\boldsymbol{\sigma}_{n,t}=(s_{n,1,t},\ldots,s_{n,V,t})
\end{equation}
forms a semantic signature, where $s_{n,v,t}$ denotes either a view-local mask ID or a visibility state in camera $v$.
Sufficiently supported signatures propose cross-view object hypotheses.
Spatially adjacent hypotheses sharing at least one view-local ID are merged into candidate \emph{identity backbones}, accommodating mask fragmentation and missing view components.
Visibility-aware voting distinguishes visible, out-of-view, and occluded projections, so partially occluded evidence is downweighted rather than removed.
Because SAM3 IDs are view-local, different candidate backbones may reuse the same ID and therefore cannot all be valid.
We therefore construct a conflict graph and solve a maximum-weight independent set (MWIS), using visibility-aware voxel support as candidate weights.
The selected non-conflicting backbones assign each view-local ID to at most one global object and define the global IDs used for tracking.

The pipeline is training-free for the target 3D tracking task, requiring neither 3D box supervision nor detector training on the collected intersections.
It requires accurate calibration and sufficient overlap, but no LiDAR or layout-specific retraining.
At a new intersection, only the cached voxel projections and site-specific search region are rebuilt from its calibration, while the tracking procedure and parameters remain unchanged.

\subsubsection{Frame-Level Metric Cuboid Fitting}

After MWIS selects a consistent set of identity backbones, we optimize a metric cuboid $\mathcal B$ for each object using
\begin{equation}
J(\mathcal B)=
\sum_{n\in\mathcal B}w_n
-\lambda\frac{V_{\mathcal B}}{\Delta^3},
\end{equation}
where $w_n$ is the visibility-aware voxel support, $V_{\mathcal B}$ is cuboid volume, $\Delta$ is the voxel-grid spacing, and the volume term penalizes unsupported space.
Category-specific size and height bounds remove implausible cuboids.

\subsubsection{Track Construction and Temporal Refinement}

Each MWIS-selected backbone links the SAM3 identities of one physical object across cameras.
Because its component identities are video-consistent within their views, the resulting global identity propagates across frames even when one component temporarily disappears.
Track-level refinement treats per-frame cuboids as noisy observations: it smooths centers, estimates a consensus size, corrects unstable headings using motion, and fills short gaps caused by occlusion.
Each output track therefore contains a persistent ID and a time-indexed sequence of metric 3D boxes.

\subsection{Oracle-Grounded Structured VQA}

\begin{figure*}[!t]
    \centering
    \includegraphics[width=\textwidth]{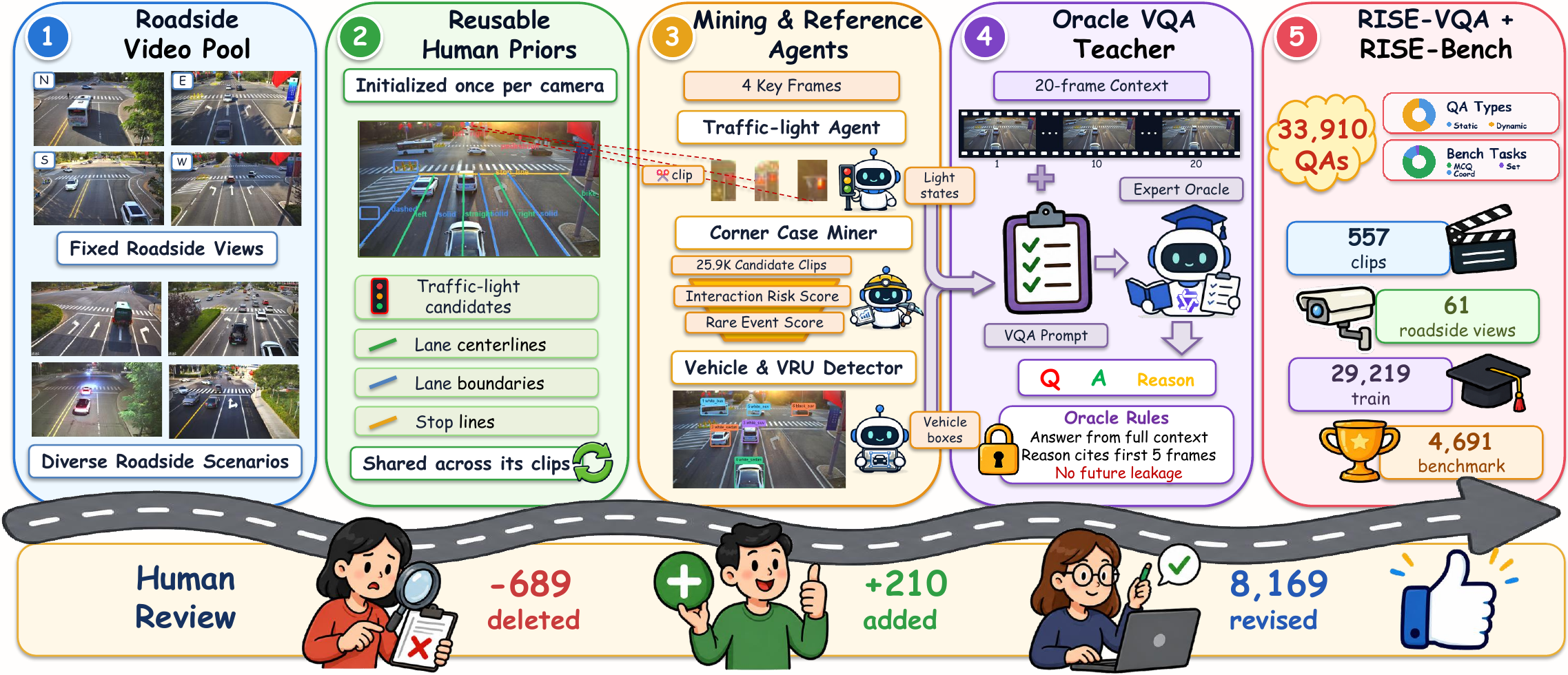}
    \caption{Structured VQA production in RISE. The pipeline mines high-value clips, constructs frame-level references, generates Oracle-grounded predictive QA pairs, and applies targeted human review.}
    \label{fig:semantic_pipeline}
\end{figure*}

\Cref{fig:semantic_pipeline} summarizes the constrained-Oracle workflow used to construct RISE-VQA.

\subsubsection{Data Collection and Scene Mining}

We collect synchronized roadside videos from 16 urban intersections, yielding 61 fixed-camera views; 13 intersections provide four directional views.
Each selected VQA clip contains 20 frames sampled at 5\,Hz.
The question and queried-agent box are anchored at $I_0$, $I_1$--$I_4$ provide observed motion context, and later key frames supply Oracle-only targets.

To avoid spending full-sequence annotation on routine traffic, a lightweight MLLM scores low-rate summaries of 25,910 candidate clips by corner-case severity and interaction risk.
Selected clips are then expanded into full 5-Hz sequences for structured QA construction.

\subsubsection{Infrastructure Priors and Structured References}

Directly asking a single MLLM to localize small traffic signals and bind road users across a long clip is unreliable.
Fixed roadside views allow us to reuse infrastructure priors across clips, including signal regions, lane geometry, and stop lines.
Specialized reference modules recognize cropped signal states and localize road users with coarse headings on selected key frames.
These structured references provide localization and association cues only to the annotation Oracle; they are never exposed to models during training or evaluation.

\subsubsection{Observation--Supervision Separation}

Question templates are restricted to evidence available in the observed prefix.
For predictive QA, the Oracle uses the complete sequence and structured references to derive realized future boxes, maneuvers, and interactions.
Training and evaluation receive only $I_0$--$I_4$ and questions anchored to observed boxes; future frames, references, and Oracle reasoning remain hidden.

\begin{table}[!t]
\centering
\small
\caption{Observation--supervision boundary in the constrained Oracle protocol. Realized future observations determine predictive targets but remain hidden from train/test inputs.}
\label{tab:oracle}
\setlength{\tabcolsep}{3pt}
\begin{tabular}{lcc}
\toprule
\textbf{Information} & \textbf{Annotation Role} & \textbf{Train/Test Input} \\
\midrule
Early frames $I_0$--$I_4$ & Evidence & Yes \\
Future frames $I_5$--$I_{19}$ & Oracle evidence & No \\
Future boxes & Supervision target & No \\
Infrastructure priors & Reference & No \\
\bottomrule
\end{tabular}
\end{table}

\subsubsection{Human Review and Visual Grounding}

Five annotators review the generated QA through add, revise, and delete operations, producing more than 8,000 edits, primarily on dynamic questions involving subtle inter-frame motion.
Each question type undergoes first-pass review and secondary checking, with uncertain cases adjudicated separately.
Human review also enforces the observation boundary by checking that question wording contains no future-derived cues.
Released QA refers to traffic agents by their observed-frame bounding boxes rather than attributes such as color or type.
Models must therefore localize the queried agent from the box and maintain its correspondence across the observed frames before answering.

\begin{figure}[!t]
    \centering
    \includegraphics[width=\columnwidth]{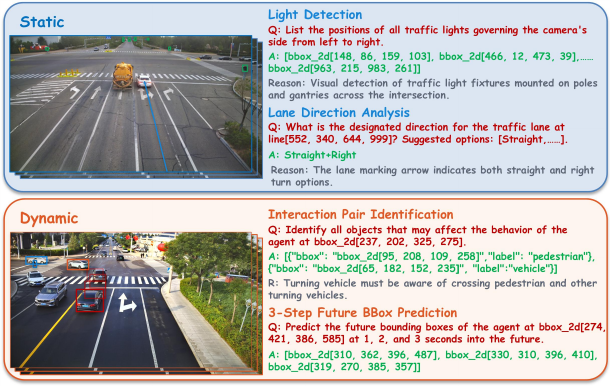}
    \caption{Examples of bbox-grounded structured VQA. Boxes identify queried agents and future locations, while colored geometry provides infrastructure and interaction context.}
    \label{fig:qa_examples}
\end{figure}

\section{Experiments}

\subsection{Dataset and RISE-Bench}

\begin{figure*}[!t]
\centering
\includegraphics[width=\textwidth]{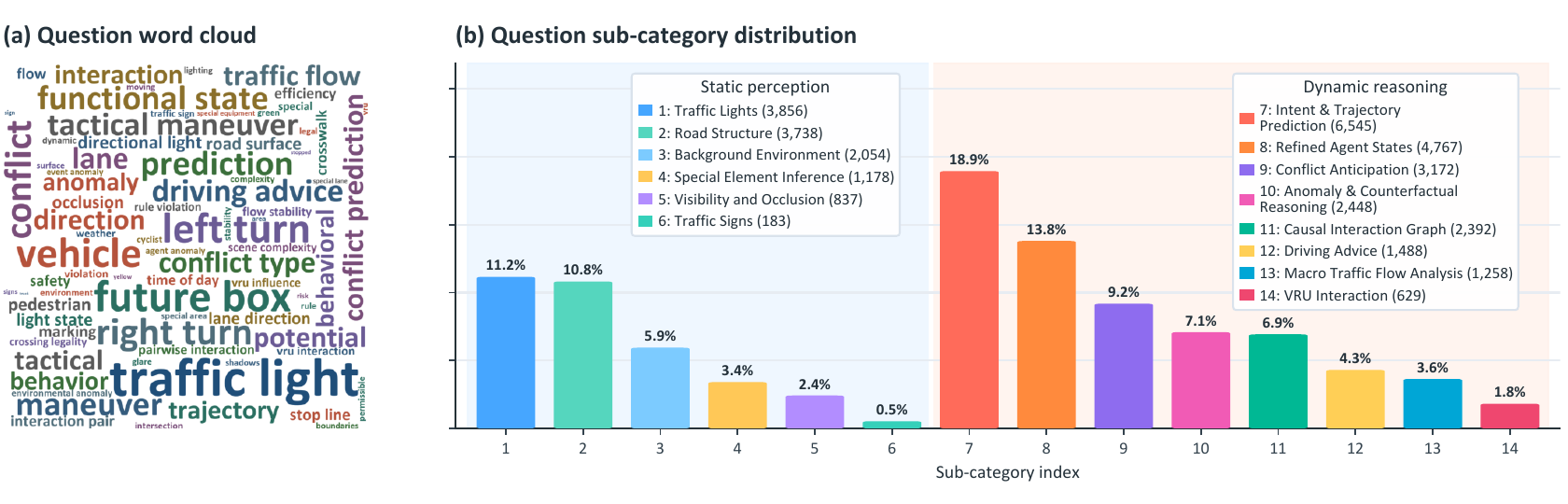}
\caption{Distribution of the 34,545 human-reviewed QA pairs before benchmark filtering. Left: frequent concepts in the question corpus. Right: fine-grained sub-category counts.}
\label{fig:qa_distribution}
\end{figure*}

The reviewed corpus contains 34,545 QA pairs, including 11,846 static-environment and 22,699 dynamic-behavior questions.
We hold out three complete intersections, covering eleven roadside views and both cross- and T-junction layouts.
By separating complete intersections rather than clips, this protocol avoids shared layout and camera context and measures cross-intersection generalization~\cite{sekaran2025urbaning}, as validated in \Cref{tab:split_ablation}.
From the 5,326 held-out instances, RISE-Bench retains 4,691 with deterministic task-specific metrics and excludes 635 open-ended or weakly structured instances.
Together with 29,219 training instances, the released dataset contains 33,910 QA pairs.

\begin{table}[!t]
\centering
\small
\caption{RISE-VQA and RISE-Bench statistics.}
\label{tab:dataset}
\setlength{\tabcolsep}{1.5pt}
\begin{tabular*}{\columnwidth}{@{\extracolsep{\fill}}lclc@{}}
\toprule
\textbf{Metric} & \textbf{Value} & \textbf{Metric} & \textbf{Value} \\
\midrule
Intersections & 16 & Roadside views & 61 \\
4-view / 3-view & 13 / 3 & Event clips & 557 \\
Reviewed QA & 34,545 & Released QA & 33,910 \\
Training QA & 29,219 & Held-out Pool & 5,326 \\
RISE-Bench QA & 4,691 & Reviewed Static & 11,846 (34.3\%) \\
& & Reviewed Dynamic & 22,699 (65.7\%) \\
\bottomrule
\end{tabular*}
\end{table}

RISE-Bench has three task families: multiple-choice questions cover semantic traffic understanding; coordinate questions cover infrastructure geometry and future boxes; and set-valued questions identify interacting agents or pairs.

\begin{table}[!t]
\centering
\small
\caption{RISE-Bench composition.}
\label{tab:benchmark}
\begin{tabular}{lcc}
\toprule
\textbf{Task Family} & \textbf{\#QA} & \textbf{Metric} \\
\midrule
Multiple choice & 3,794 & Choice score \\
Coordinate output & 745 & IoU / F1 / C-ADE \\
Set prediction & 152 & Precision / Recall / F1 \\
\midrule
Total & 4,691 & -- \\
\bottomrule
\end{tabular}
\end{table}

\subsection{Models and Metrics}

We evaluate three open-source VLMs---Qwen2.5-VL-7B, InternVL3-8B, and MiniCPM-V-4.5-8B---under zero-shot (ZS) and fine-tuned (FT) settings.
FT denotes five-frame LoRA adaptation, and both training and evaluation observe only the first five frames of each clip.
LoRA experiments use LLaMA-Factory~\cite{zheng2024llamafactory} with rank 8, scaling factor 16, learning rate $5\times10^{-5}$, effective batch size 32, and three epochs on 29,219 training instances; the vision encoder and multimodal projector remain frozen.
Primary runs use direct-answer inference with thinking disabled, while thinking-enabled proprietary models are included as strong zero-shot references marked by $^\dagger$.
The ZS--FT comparison measures the effect of domain-specific adaptation, and the 1f--5f ablation measures the contribution of temporal context.

Choice score assigns full, partial, or zero credit using predefined mappings for each question type; partial credit applies only to ordinal or adjacent choices.
Det-F1 matches predicted and reference boxes by IoU, while Line-F1 matches lane and stop-line segments by endpoint distance.
Future boxes are evaluated by trajectory IoU (T-IoU), which averages box overlap across future steps, and center average displacement error (C-ADE),
\begin{equation}
\mathrm{C\text{-}ADE}=\frac{1}{3N}\sum_{i=1}^{N}\sum_{\tau=1}^{3}
\lVert\hat{\mathbf c}_{i,\tau}-\mathbf c_{i,\tau}\rVert_2.
\end{equation}
Here $\mathbf c_{i,\tau}$ is the ground-truth box center in the common $1000\times1000$ relative coordinate space.
Interaction predictions use label-agnostic bbox set matching, and Inter-F1 is macro-averaged over QA samples.

\subsection{Structured Vision-Language Results}

\begin{table*}[!t]
\centering
\small
\caption{Main results on RISE-Bench. S-MCQ and D-MCQ denote static and dynamic MCQ scores. ZS-5f and FT-5f denote five-frame zero-shot inference and LoRA adaptation, respectively. $^\ddagger$ marks bbox evaluation normalized from model-native serialization and coordinate systems; $^\dagger$ marks thinking-enabled proprietary zero-shot runs. Other rows use direct-answer inference. Bold indicates the best result within the proprietary-reference or fine-tuned group.}
\label{tab:main_results}
\setlength{\tabcolsep}{2.75pt}
\begin{tabular}{@{}lll|ccc|cc|cc|c@{}}
\toprule
\textbf{Model} & \textbf{Access} & \textbf{Setting} & \textbf{S-MCQ} & \textbf{D-MCQ} & \textbf{MCQ} & \textbf{Det-F1} & \textbf{Line-F1} & \textbf{T-IoU} & \textbf{C-ADE$\downarrow$} & \textbf{Inter-F1} \\
\midrule
Qwen2.5-VL-7B & Open & ZS-5f$^\ddagger$ & .479 & .533 & .510 & .078 & .001 & .305 & 120.4 & .066 \\
InternVL3-8B & Open & ZS-5f$^\ddagger$ & .503 & .424 & .458 & .029 & .029 & .050 & 306.8 & .015 \\
MiniCPM-V-4.5-8B & Open & ZS-5f$^\ddagger$ & .507 & .571 & .544 & .061 & .055 & .257 & 131.3 & .042 \\
\midrule
GPT-5.5 & Closed & ZS-5f & .558 & .593 & .578 & .392 & .798 & .248 & 117.8 & .106 \\
GPT-5.5$^\dagger$ & Closed & ZS-5f & .610 & .632 & .623 & .649 & \textbf{.920} & .402 & \textbf{60.8} & .238 \\
Gemini-3.1-Pro$^\dagger$ & Closed & ZS-5f$^\ddagger$ & .502 & .491 & .495 & .187 & .719 & .323 & 112.9 & .214 \\
Qwen3.7-Plus & Closed & ZS-5f & .635 & .649 & .643 & .382 & .738 & .311 & 99.1 & .251 \\
Qwen3.7-Plus$^\dagger$ & Closed & ZS-5f & .640 & .666 & .655 & .599 & .791 & .374 & 64.2 & .332 \\
\midrule
Qwen2.5-VL-7B & Open & FT-5f & \textbf{.782} & .781 & \textbf{.781} & .643 & \textbf{.860} & .381 & 74.8 & .271 \\
InternVL3-8B & Open & FT-5f & .734 & \textbf{.791} & .767 & \textbf{.672} & .845 & \textbf{.433} & \textbf{62.2} & \textbf{.350} \\
MiniCPM-V-4.5-8B & Open & FT-5f & .722 & .764 & .746 & .648 & .771 & .344 & 91.2 & .305 \\
\bottomrule
\end{tabular}
\end{table*}

\Cref{tab:main_results} shows that domain-specific adaptation consistently improves all three open-source backbones across answer formats.
For rows marked by $^\ddagger$, native bbox serialization and coordinates are normalized before scoring, so the gains cannot be attributed solely to output-format alignment.
Among the fine-tuned models, Qwen2.5-VL-7B leads in aggregate MCQ and Line-F1, whereas InternVL3-8B performs best on dynamic MCQ, detection, future-box, and interaction metrics.
This variation supports separately evaluating semantic choices, geometric grounding, future localization, and interaction sets.
Thinking improves both paired proprietary baselines but does not dominate every structured task.

\subsection{Temporal Context Ablation}

\begin{table}[!t]
\centering
\normalsize
\caption{Temporal-context ablation. Each pair uses the same backbone and LoRA recipe; only the number of input frames differs.}
\label{tab:temporal}
\setlength{\tabcolsep}{1pt}
\begin{tabular*}{\columnwidth}{@{\extracolsep{\fill}}llcccc@{}}
\toprule
\textbf{Model} & \textbf{Fr.} & \textbf{S-MCQ} & \textbf{D-MCQ} & \textbf{T-IoU} & \textbf{C-ADE$\downarrow$} \\
\midrule
Qwen2.5-VL & 1f & .784 & .730 & .312 & 109.1 \\
& 5f & .782 & .781 & .381 & 74.8 \\
\midrule
InternVL3 & 1f & .698 & .738 & .324 & 104.0 \\
& 5f & .734 & .791 & .433 & 62.2 \\
\midrule
MiniCPM-V- & 1f & .727 & .730 & .322 & 115.1 \\
4.5-8B & 5f & .722 & .764 & .344 & 91.2 \\
\bottomrule
\end{tabular*}
\end{table}

Across all three backbones, 5f improves D-MCQ and T-IoU while reducing C-ADE, providing consistent evidence that observed motion benefits dynamic reasoning and future localization.
Because static questions are anchored in the first frame shared by both settings, S-MCQ provides a useful reference: it changes little for Qwen2.5-VL and MiniCPM-V, while improving modestly for InternVL3.
The gain remains backbone-dependent, with InternVL3 showing the largest trajectory improvement.

\begin{table}[!t]
\centering
\small
\caption{Split-policy ablation with Qwen2.5-VL-7B. Held and Clip denote intersection-held-out and clip-random train/test splitting, respectively. All denotes their respective test sets (4,691 and 4,359 QA pairs); Com.\ denotes the same 1,451 QA pairs absent from both training sets.}
\label{tab:split_ablation}
\setlength{\tabcolsep}{0pt}
\begin{tabular*}{\columnwidth}{@{\extracolsep{\fill}}llcccccc@{}}
\toprule
\textbf{Eval.} & \textbf{Split} & \textbf{S-MCQ} & \textbf{D-MCQ} & \textbf{Det-F1} & \textbf{Line-F1} & \textbf{T-IoU} & \textbf{C-ADE$\downarrow$} \\
\midrule
All & Held & .782 & .781 & .643 & .860 & .381 & 74.8 \\
All & Clip & .857 & .806 & .754 & .944 & .379 & 62.0 \\
\midrule
Com. & Held & .772 & .775 & .574 & .846 & .399 & 71.0 \\
Com. & Clip & .841 & .796 & .654 & .896 & .420 & 57.0 \\
\bottomrule
\end{tabular*}
\end{table}

We retrain Qwen2.5-VL-7B under the intersection-held-out and clip-random train/test partitions.
Because the full test sets differ in composition, their rows provide only an overall view.
On the common 1,451-QA subset, clip-random training is stronger across all metrics, including S-MCQ (.841 vs.\ .772), D-MCQ (.796 vs.\ .775), and Det-F1 (.654 vs.\ .574).
Shared intersection context therefore makes clip-level splitting easier, motivating intersection-held-out evaluation.

\subsection{3D Tracking Quality Analysis}

\begin{figure}[!t]
    \centering
    \includegraphics[width=\columnwidth]{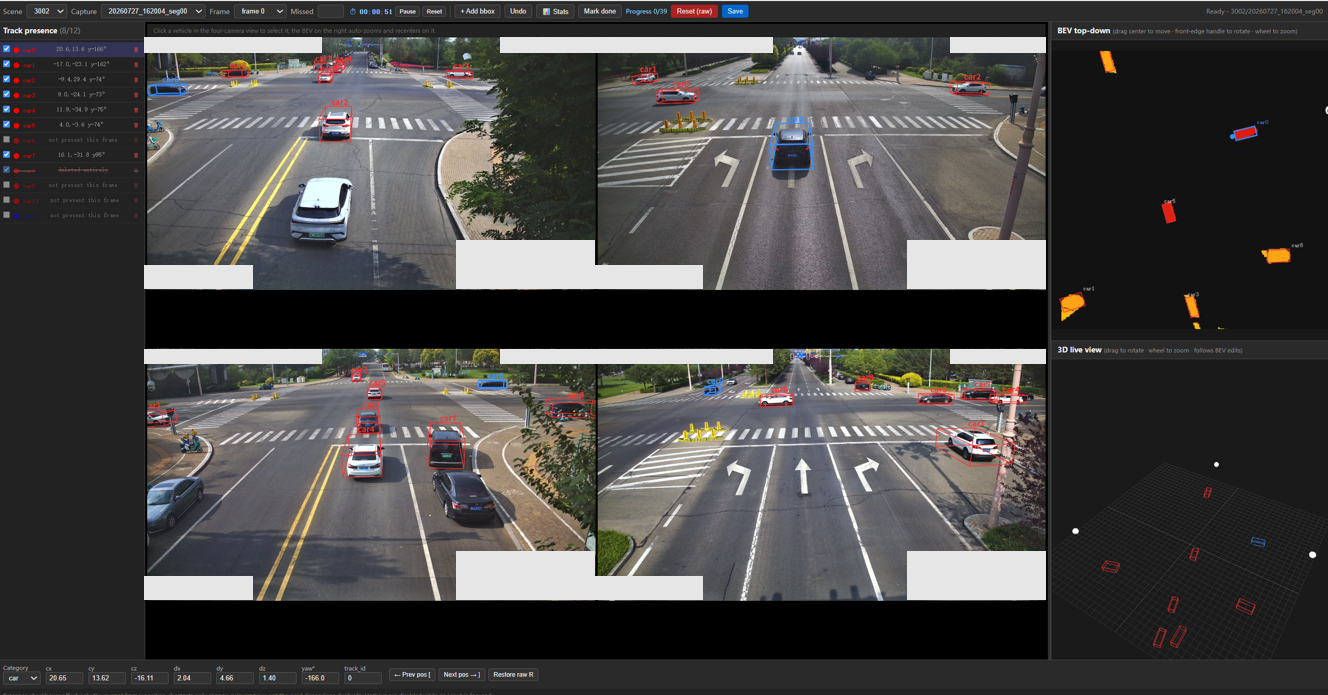}
    \caption{Human review interface for initialized 3D tracks. Multi-view projections and the semantic-voxel BEV guide manual correction.}
    \label{fig:bbox_review_interface}
\end{figure}

We evaluate the unedited tracker output as the prediction and the manually corrected tracks as GT.
As shown in \Cref{fig:bbox_review_interface}, the tracker provides initial 3D tracks, while synchronized views and the semantic-voxel BEV help annotators add missed eligible vehicles, remove false tracks, and correct box geometry and identity.

All evaluated scenes use the same voxelized world-coordinate region, $x,y\in[-40,40]$\,m, with an approximately $0.2$\,m grid step.
We regard a vehicle as observable when it is clearly identifiable at sufficient scale in at least one camera; tiny or distant vehicles are excluded.
Among observable vehicles, those inside the voxel region whose projected regions lie within the image bounds of at least three synchronized cameras enter the frame-level tracking GT, regardless of occlusion.
The remainder are recorded as scope exclusions rather than tracker false negatives; tracker false negatives are unmatched eligible GT boxes.
For each 10-second clip, annotators review 20 frames sampled at 2\,Hz.
To keep manual correction tractable, the current evaluation uses 20 sampled clips from six intersections within the fixed ROI and three-view coverage scope.
All intersections use the same tracking parameters; only calibration-dependent voxel projections and site-specific search regions are reconfigured.
This clip-based protocol limits the scale of the present evaluation rather than the temporal duration supported by the tracker.

\begin{table}[!t]
    \centering
    \small
    \caption{Statistics of the human-reviewed subset. Dimensions are adjusted once per track; a frame requires no frame-wise correction when its XY center and heading remain unchanged.}
    \label{tab:preliminary_3d_statistics}
    \setlength{\tabcolsep}{2.6pt}
    \begin{tabular*}{\columnwidth}{@{\extracolsep{\fill}}lclc@{}}
        \toprule
        \textbf{Metric} & \textbf{Value} & \textbf{Metric} & \textbf{Value} \\
        \midrule
        Generated boxes & 3,240 & GT boxes & 3,317 \\
        GT tracks & 251 & Scope-excl.\ inst. & 1,158 \\
        Exclusion rate & 25.9\% & No frame-wise edit & 1,900 (57.3\%) \\
        \bottomrule
    \end{tabular*}
\end{table}

Prediction--GT association is determined geometrically rather than by inherited track IDs, allowing detection errors, localization errors, and identity consistency to be evaluated separately.
Specifically, we perform class-aware frame-wise Hungarian matching using oriented BEV IoU; matches with BEV IoU below 0.5 are rejected, and the remaining unmatched predictions and eligible GT boxes are counted as FP and FN, respectively.

\begin{table}[!t]
    \centering
    \small
    \caption{3D tracking quality at a BEV IoU matching threshold of 0.5. MOTP is the mean BEV IoU over valid matches.}
    \label{tab:preliminary_3d_quality}
    \setlength{\tabcolsep}{2.6pt}
    \begin{tabular*}{\columnwidth}{@{\extracolsep{\fill}}lclc@{}}
        \toprule
        \textbf{Metric} & \textbf{Value} & \textbf{Metric} & \textbf{Value} \\
        \midrule
        MOTA $\uparrow$ & 66.9 & MOTP$_{\mathrm{BEV}}$ $\uparrow$ & 87.0 \\
        IDS $\downarrow$ & 40 & F1 $\uparrow$ & 83.9 \\
        Precision / Recall $\uparrow$ & 84.9 / 82.9 & XY err. (m) $\downarrow$ & .071 \\
        \bottomrule
    \end{tabular*}
\end{table}

\begin{figure}[!t]
    \centering
    \includegraphics[width=\columnwidth]{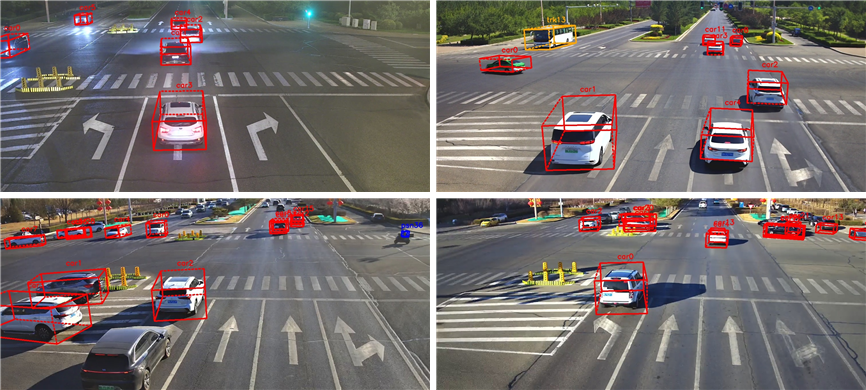}
    \caption{Projected cuboids from the multi-view 3D tracker across representative roadside sequences.}
    \label{fig:track_examples}
\end{figure}

\Cref{fig:track_examples} shows representative projected cuboids across roadside sequences.
The peripheral vehicle omitted near an overlap boundary falls outside the predefined three-view evaluation scope rather than being counted as a tracker false negative.

\section{Conclusion}

We presented RISE, a framework for roadside sequence understanding through two complementary tasks: metric 3D tracking and structured vision-language reasoning.
By exploiting complementary spatial and temporal evidence, RISE supports persistent 3D tracking and bbox-grounded prediction without exposing future evidence to evaluated models.
Human-corrected evaluation on 20 clips yields 66.9 MOTA for the generated tracks within the defined multi-view scope.
Experiments on RISE-Bench demonstrate clear benefits from domain adaptation and generally from temporal context, while revealing remaining challenges in spatial grounding, future localization, and interaction reasoning.
These findings highlight both the value and the current limitations of sequence-level roadside understanding.

Several limitations remain.
Image-only 3D tracking depends on accurate calibration, reliable segmentation, and sufficient camera overlap, with weaker geometric support under heavy occlusion and near shared-view boundaries.
On the vision-language side, the current 3.8-second clips primarily assess near-term reasoning from individual roadside views.
Extending the observation horizon without sacrificing temporal resolution, together with coordinated reasoning across synchronized views, would enable the benchmark to capture longer and more complex traffic evolution.
A promising direction is to use persistent metric 3D trajectories as explicit spatiotemporal grounding for vision-language models, enabling more reliable reasoning about long-term motion and multi-agent interactions.

\bibliography{01_roadside_perception,02_llm_annotation,03_3d_detection,04_scene_mining}

\end{document}